\documentclass[letterpaper,10pt,conference]{IEEEtran}

\usepackage[english]{babel}

\usepackage{cite}

\usepackage[hidelinks]{hyperref}
\usepackage{url}
\newcommand\rurl[1]{%
  \href{http://#1}{\nolinkurl{#1}}%
}

\usepackage{newtxmath}
\usepackage[binary-units]{siunitx}
\usepackage[acronym]{glossaries}
\glsdisablehyper
\newacronym{sdk}{SDK}{Software Development Kit}
\newacronym{vo}{VO}{Visual Odometry}

\usepackage{cleveref}
\crefname{table}{Table}{Tables}
\crefname{figure}{Figure}{Figures}
\crefname{section}{Section}{Sections}
\usepackage{multicol}
\usepackage{multirow}

\usepackage{graphicx}

\usepackage{cuted}
\usepackage{listings}

\usepackage{copyright}

\usepackage[labelformat=simple]{subcaption}

\usepackage{xcolor}

\newcommand{\review}[1] {#1}

\providecommand{\keywords}[1]
{
  \small	
  \textbf{\textit{Keywords---}} #1
}

\newcommand\datasetName{The Oxford Road Boundaries Dataset}
\newcommand\totalCurated{47639}
\newcommand\totalUncurated{62605}

\begin{document}

\title{
\Large \bf
\datasetName
}
\author{
Tarlan Suleymanov, Matthew Gadd, Daniele De Martini, and Paul Newman\\
Oxford Robotics Institute, University of Oxford, United Kingdom\,\\
\texttt{\{tarlan,mattgadd,daniele,pnewman\}@robots.ox.ac.uk}
}
\maketitle
\copyrightnotice

\begin{strip}\centering
\includegraphics[width=0.245\textwidth]{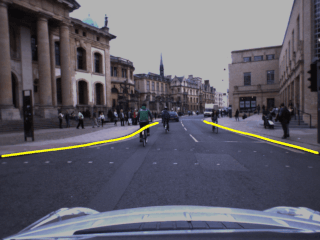} \hskip -0.5ex
\includegraphics[width=0.245\textwidth]{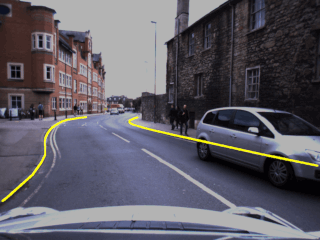} \hskip -0.5ex
\includegraphics[width=0.245\textwidth]{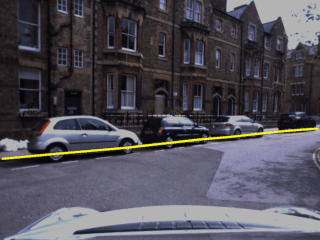} \hskip -0.5ex
\includegraphics[width=0.245\textwidth]{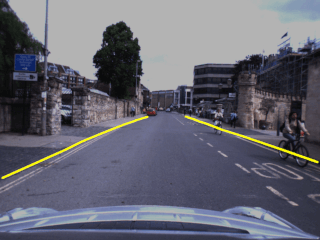}
\vskip 0.3ex
\includegraphics[width=0.245\textwidth]{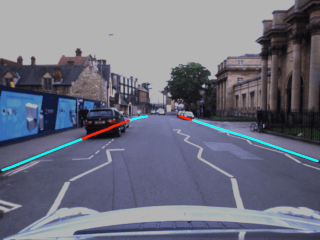} \hskip -0.5ex
\includegraphics[width=0.245\textwidth]{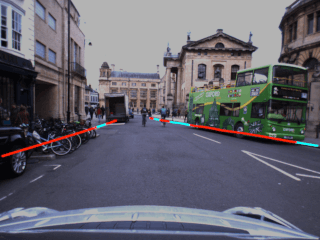} \hskip -0.5ex
\includegraphics[width=0.245\textwidth]{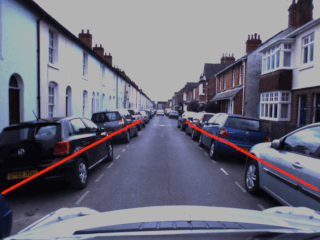} \hskip -0.5ex
\includegraphics[width=0.245\textwidth]{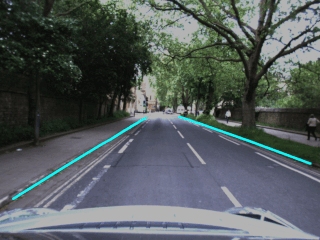}
\captionof{figure}{Dataset examples. First row: ``raw''. Second row: ``classified'' or ``partitioned'' -- visible (cyan) and occluded (red). First and third columns are left lenses, second and fourth columns are right lenses. For these and other visualisations, these road boundaries are dilated, but are in fact one pixel in width in the released data.
\label{fig:main_figure}}
\end{strip}

\begin{abstract}
In this paper we present \textit{\datasetName}, designed for training and testing machine-learning-based road-boundary detection and inference approaches.
We have hand-annotated two of the \SI{10}{\kilo\metre}-long forays from the \textit{Oxford Robotcar Dataset} and generated from other forays several thousand further examples with semi-annotated road-boundary masks.
To boost the number of training samples in this way, we used a vision-based localiser to project labels from the annotated datasets to other traversals at different times and weather conditions.
As a result, we release \num{\totalUncurated} labelled samples, of which \num{\totalCurated} samples are curated.
Each of these samples contain both \textit{raw} and \textit{classified} masks for left and right lenses.
Our data contains images from a diverse set of scenarios such as straight roads, parked cars, junctions, etc.
Files for download and tools for manipulating the labelled data are available at: \urlstyle{tt}\rurl{oxford-robotics-institute.github.io/road-boundaries-dataset}
\end{abstract}
\keywords{road boundary, curb, kerb, dataset}

\section{Introduction}%
\label{sec:introduction}

\begin{figure}[h]
\centering
\includegraphics[width=0.49\columnwidth]{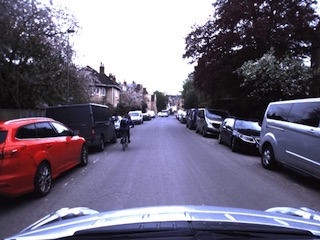} \hskip -0.5ex
\includegraphics[width=0.49\columnwidth]{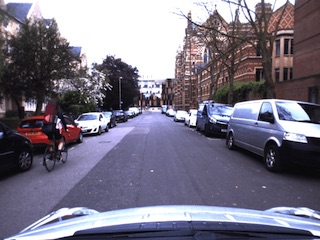} \vskip 0.3ex
\includegraphics[width=0.49\columnwidth]{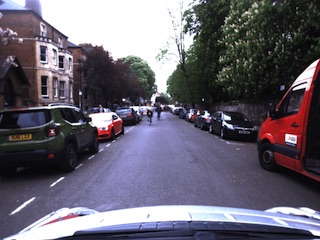} \hskip -0.5ex
\includegraphics[width=0.49\columnwidth]{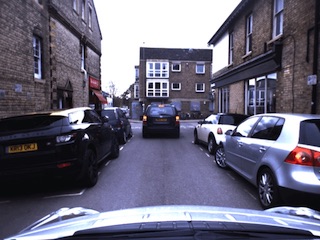}
\caption{Examples where road boundaries are fully occluded.}
\label{fig:occluded_roads_examples}
\vspace{-.5cm}
\end{figure}

Obtaining well-generalising, high-performance deep networks typically requires large quantities of training samples.
Even if easy to hand-annotate -- and this is often not the case -- this is an inordinate workload for the researcher.
To cope with the variability of \emph{road boundaries}, the required data should equally incorporate great variability changes in environment, scale, appearance, colour, background clutter, occlusion, perspective and illumination.
Yet, fine-grained annotation of data requires time-consuming human effort where labels of different classes must be assigned to outlined distinct regions.
Nevertheless, finding true road boundaries is crucial for autonomous vehicles as they legally and intentionally delimit drive-able surfaces.
Indeed, as somewhat durable parts of the road infrastructure, lane markings, traffic signs, feature points or road boundaries are often used as features for localisation~\cite{sun20193d}.
Moreover, detecting only visible road boundaries is often not sufficient in urban scenarios -- examples in \cref{fig:occluded_roads_examples} show road boundaries either partially or fully occluded.
Indeed, there is often no visible road boundary to detect, as in~\cref{fig:occluded_roads_examples}.
In these sorts of scenarios, the vehicle should still sufficiently sense the environment to move safely about it.
For this reason, datasets incorporating road boundaries should include a large number of samples in order to capture the variability of road boundaries.
They should also contain annotations for both visible and occluded road boundaries.
To this end, we have developed a framework for efficient annotations within a reasonable amount of time.
In this paper, we present \textit{\datasetName}, which we generated using this framework.

\section{Related Work}
\label{sec:related_work}

Road-boundary detection is useful for autonomous driving in that it can facilitate safe driving in a legally demarcated area.
Work in this area includes~\cite{wijesoma2004road,chen2017rbnet,liang2019convolutional,perng2020development}.

Autonomous driving datasets abound~\cite{geiger2013vision,cordts2016cityscapes,wang2019apolloscape,caesar2019nuscenes,maddern20171,barnes2020rrcd,carlevaris2016university,geyer2020a2d2} and accurate, human-derived annotations for a wide number of semantic classes are available in these releases.
However, road boundaries are not included, regardless of their importance for scene understanding, lane-level localisation, mapping and planning in autonomous driving.

The work which is mostly related to ours is~\cite{xu2021topo}; yet, since it is based on aerial imagery, it does not provide synchronisation to vehicle-mounted sensors -- and is not released for download at the time of writing.

Works based on the proposed dataset include~\cite{Suleymanov2018Curb,2018ITSC_kunze,suleymanov2019thesis}, where CNN-based deep-leaning models -- \textit{Visible Road Boundary Detection (VRBD)} and \textit{Occluded Road Boundary Inference (ORBI)} -- were proposed for detection and inference of visible and occluded road boundaries and were applied for scene understanding and lateral localisation problems.

\begin{figure*}[!h]
    \centering
    \includegraphics[width=0.75\textwidth]{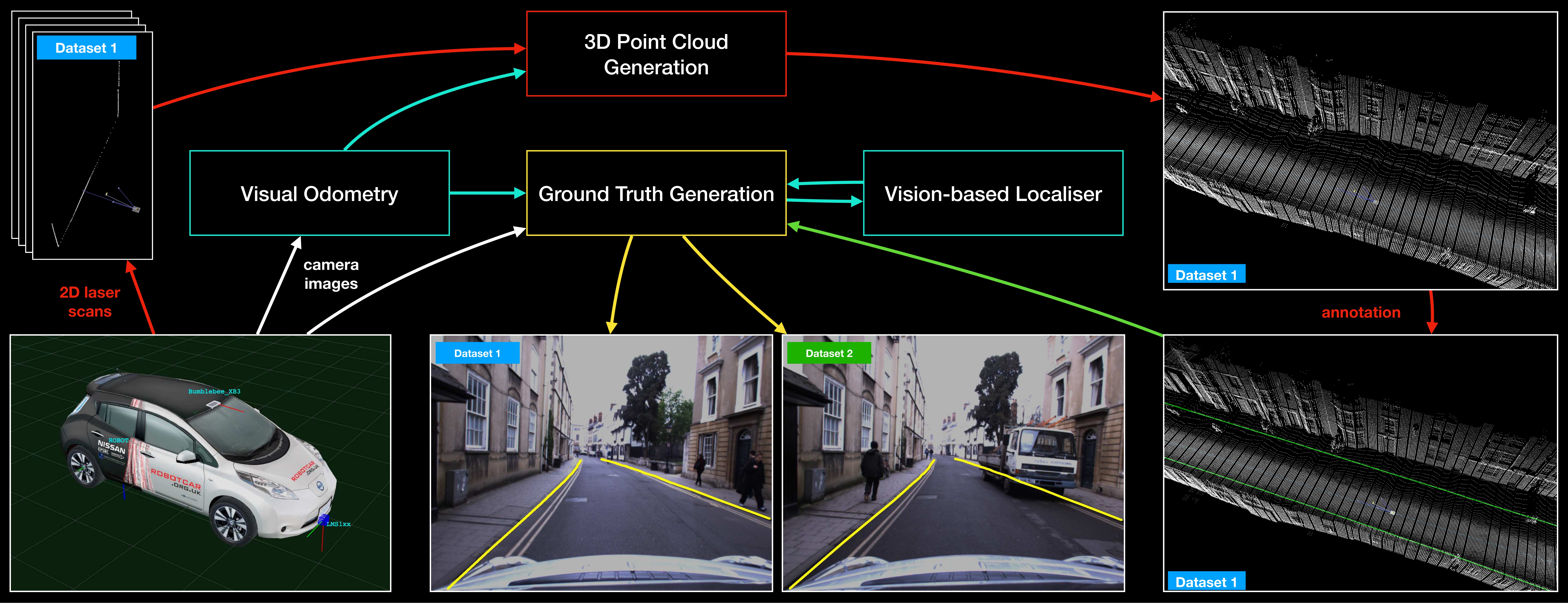}
    \caption{Road Boundary Data Annotation Framework: an efficient way of annotating road boundaries to obtain ground truth samples.}
    \label{fig:data_annotation_framework}
\end{figure*}

\section{\datasetName}

This section describes the vehicle, collection regime, annotation procedures and dataset partitioning.

The overall pipeline of the proposed road boundary data generation framework is shown in \cref{fig:data_annotation_framework}.
We generate camera-based annotation from the mutimodal collected data (\cref{sec:dataset:collection}) by projecting laser-based annotations to the stereo, forward-facing camera (\cref{sec:dataset:annotation}) through \gls{vo} interpolation.
Visual localisation tools are used to refer ground-truth annotations to other traversals for fast extension of the labels (\cref{sec:dataset:localiser}).
Manual curation is carried out to check the created labels (\cref{sec:dataset:curation}) and a deep-learning pipeline is used to automatically classify the boundaries as either visible or occluded (\cref{sec:dataset:partitioning}).

\subsection{Data Collection}
\label{sec:dataset:collection}

\begin{figure}[h]
\centering
\includegraphics[width=\columnwidth]{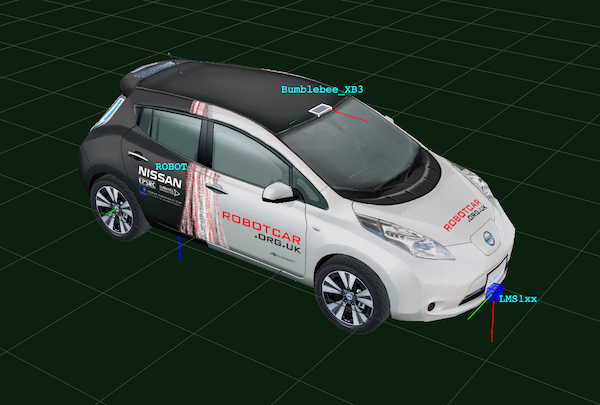}
\caption{The platform -- a modified Nissan LEAF -- used for data collection in the Oxford Road Boundary Dataset and its sensor configuration. Note, only sensors that are relevant to this dataset are shown: a forward facing Point Grey Bumblebee XB3 colour stereo camera and a SICK LMS-151 2D LiDAR vertically attached to the front of the car.}
\label{fig:robotcar}
\end{figure}

This release is an annotative exercise atop the original \textit{Oxford RobotCar Dataset}, of which we used seven forays from which approximately \num{62000} pairs of samples were generated.
\Cref{tab:dataset} summarises the forays driven.

The dataset was collected using the Oxford RobotCar platform, as in~\cite{RobotCarDatasetIJRR,barnes2020rrcd}, an autonomous-capable Nissan LEAF (\cref{fig:robotcar}).
For the purpose of data annotation (see \cref{sec:dataset:annotation}), the following sensors are of interest:
\begin{enumerate}
    \item $1 \times$ SICK LMS-151 2D LiDAR, \SI{270}{\degree} FoV, \SI{50}{\hertz}, \SI{50}{\metre} range, \SI{0.5}{\degree} resolution.
    \item $1 \times$ Point Grey Bumblebee XB3 (BBX3-13S2C38) trinocular stereo camera, $1280{\times}960{\times}3$, \SI{16}{\hertz}, $1/3''$ Sony ICX445 CCD, global shutter, \SI{3.8}{\milli\metre} lens, \SI{66}{\degree} HFoV, \SI{12/24}{\centi\metre} baseline.
\end{enumerate}
Namely, the LiDAR is used in the annotation process to project visible and occluded road boundaries into each of the camera lens's frames.

\begin{table*}[!h]
\renewcommand{\arraystretch}{1.8}
    \centering
    \begin{tabular}{c|c|c|c|c|c|c|c}
     & \multicolumn{2}{c|}{Number of frames} 
     & \multicolumn{2}{c|}{\% of occluded labels}
     & \multicolumn{2}{c|}{Download size}
     & \\
    \multirow{-2}{*}{Foray} & All & Curated & All & Curated & All & Curated & \multirow{-2}{*}{Annotation} \\
    \hline
    2015-05-26-13-59-22 & 22775 & 14890 & 31\% & 29\% & 63 GB & 41 GB & manual \\
    \hline
    2015-03-17-11-08-44 & 4869 & 3923 & 30\% & 30\% & 14 GB & 10 GB & automatic \\
    \hline
    2015-05-08-10-33-09 & 5519 & 4617 & 31\% & 31\% & 16 GB & 13 GB & automatic \\
    \hline
    2015-05-19-14-06-38 & 5424 & 4271 & 45\% & 44\% & 15 GB & 12 GB & automatic \\
    \hline
    2018-04-30-14-43-54 & 15153 & 13353 & 28\% & 29\% & 44 GB & 38 GB & manual \\
    \hline
    2019-01-10-11-46-21 & 4440 & 3392 & 39\% & 39\% & 12 GB & 10 GB & automatic \\
    \hline
    2019-01-10-12-32-52 & 4425 & 3193 & 40\% & 38\% & 12 GB & 8 GB & automatic \\
    \hline
    \textbf{Total} & \textbf{\totalUncurated} & \textbf{\totalCurated} & \textbf{33\%} & \textbf{32\%} & \textbf{176 GB} & \textbf{132 GB} & N$\backslash$A \\
    \end{tabular}
\caption{Summary of data released. \review{Manual: labels are obtained by projecting road boundaries from hand-annotated laser point cloud to images. Automatic: labels are obtained by projecting road boundaries from the hand-annotated datasets to other traversals using vision-based localiser.}}
\label{tab:dataset}
\end{table*}

\begin{figure*}[h]
    \centering
    \includegraphics[width=0.69\textwidth]{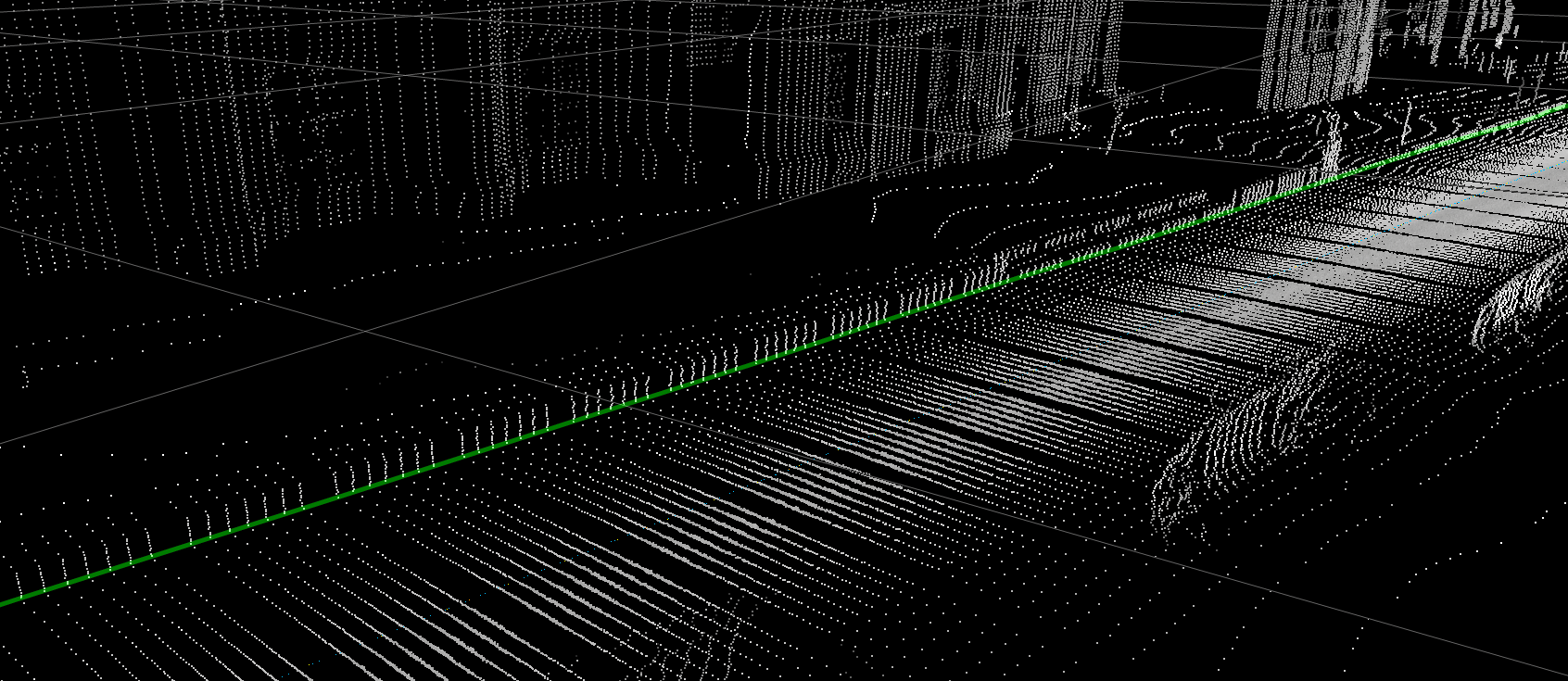} \hskip -0.5ex
    \includegraphics[width=0.30\textwidth]{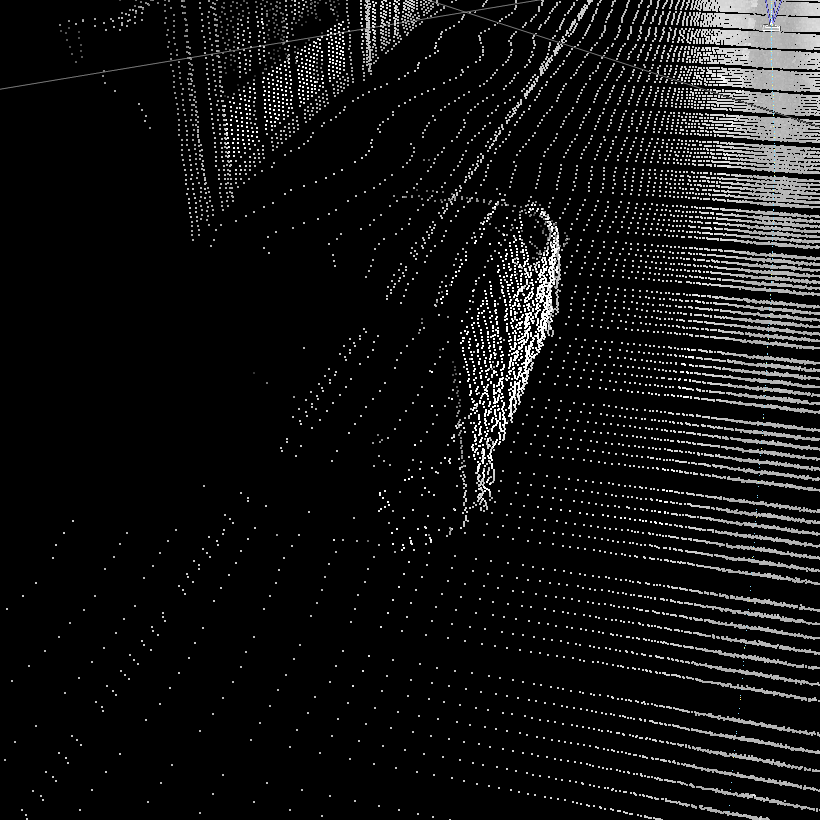}
    \caption{Left: lines are drawn between consecutive points with the same ID to annotate road boundary segments between the points. Right: laser light pulses reach the road boundary behind the parked car as they pass under the car.}
    \label{fig:annotated_road_boundary_in_pc}
\end{figure*}

\subsection{Annotation}
\label{sec:dataset:annotation}

\begin{figure}
    \centering
    \includegraphics[width=\columnwidth]{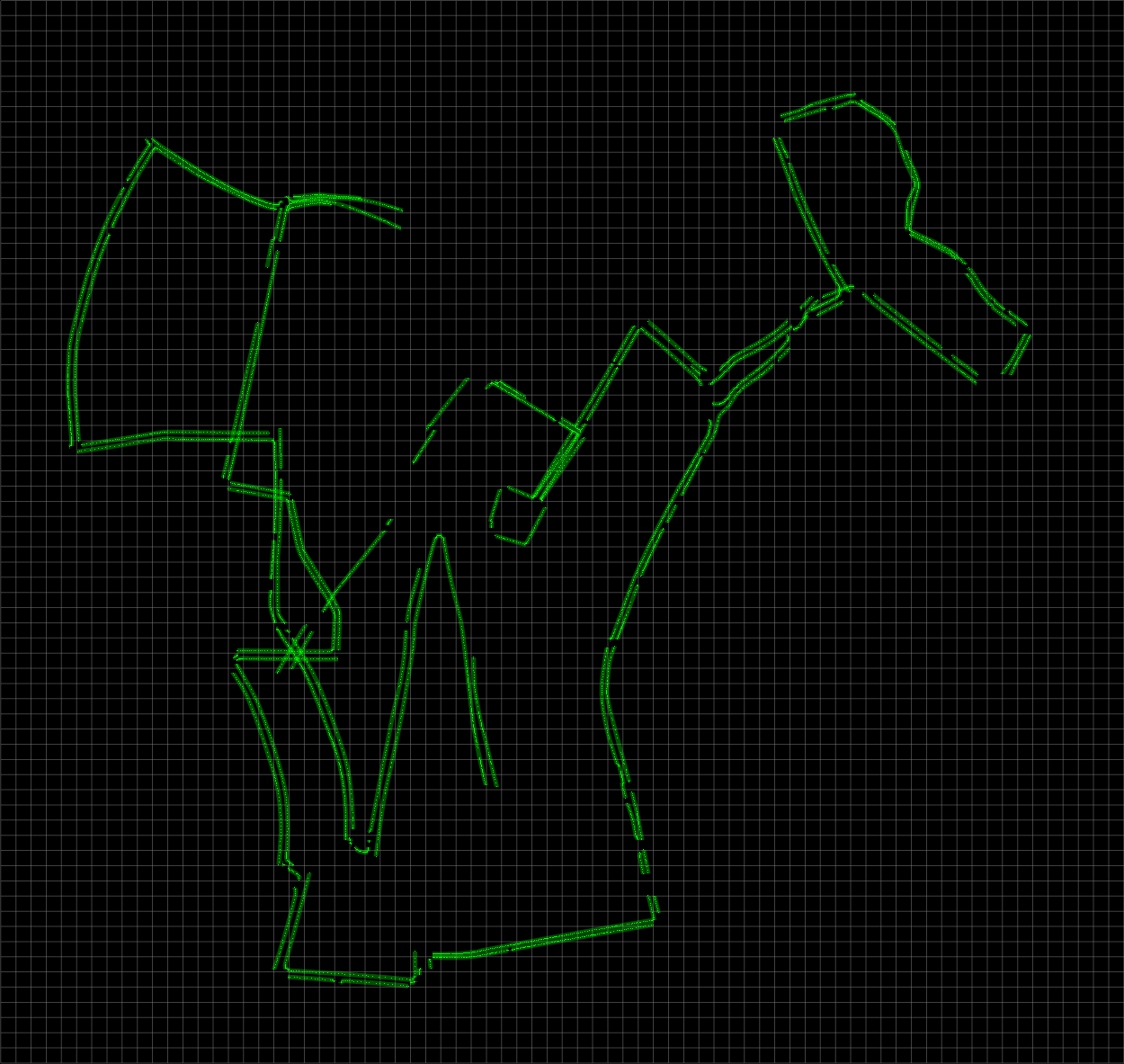}
    \caption{Bird's-eye view of the annotations of the \textit{2015-05-26-13-59-22} dataset, 10 kilometres were annotated.}
    \label{fig:annotated_dataset}
\end{figure}

Fine-grained hand-annotation of road boundaries from images would be a very time-consuming process and it would be impossible to exactly annotate the position of occluded road boundaries from the images themselves.
To tackle this problem,, we annotated the 3D pointcloud data collected by the 2D laser: as shown in \cref{fig:robotcar}, the laser is attached vertically to the front of the vehicle; as such, the light pulses not only hit the road boundaries almost perpendicularly -- which made the road boundaries easily distinguishable for annotation -- but also reach road boundaries behind parked vehicles as the pulses can pass under the vehicles  (see \cref{fig:annotated_road_boundary_in_pc}).

3D pointclouds are constructed by accumulating the push-broom returns along a pose-chain formed by \gls{vo} which is only required to be accurate in the local region around the example to be annotated -- drift is not a significant problem.
Specifically, to obtain a 3D pointcloud from the \textit{Oxford RobotCar Dataset} from 2D laser, we integrated subsequent vertical laser scans in a coherent coordinate frame.
We used \gls{vo} to estimate the vehicle’s motion and compute the transformations between subsequent scans. 
Note that, as \gls{vo} uses camera images to calculate the vehicle’s ego-motion, it provides transformations between camera timestamps which in general differ from the laser timestamps.
For this reason, we use interpolation to obtain transformations between subsequent 2D laser scans at time frames $t_l$ and $t_l'$ as follows:
\begin{align} 
    \label{eq:chain_transformation}
    T(t_l', t_l) & = T(t_l', t_{vo}') \cdot T(t_{vo}', t_{vo}) \cdot T(t_{vo}, t_l) \\
    s.t. & \ t_{vo}' \leq t_l',  \ \ t_{vo} \geq t_l \nonumber
\end{align}
where $t_{vo}'$ and $t_{vo}$ are the closest time steps of the \gls{vo} with respect to the laser frames and where $T(t_{vo}', t_{vo})$ is defined as follows:
\begin{equation}
    T(t_{vo}', t_{vo}) = \prod_{i=t_{vo}'}^{t_{vo}-1}T(i, i+1)
\end{equation}

Once we have generated 3D point clouds of the datasets, we assign the same IDs to the points lying on the same continuous boundary.
This enabled us to connect consecutive points according their IDs and fully annotate road boundary segments between the points, as shown in \cref{fig:annotated_road_boundary_in_pc}.
A Bird's-eye view of the annotations of the \textit{2015-05-26-13-59-22} foray is shown in~\cref{fig:annotated_dataset}.

\begin{figure}[!h]
\centering
\includegraphics[width=\columnwidth]{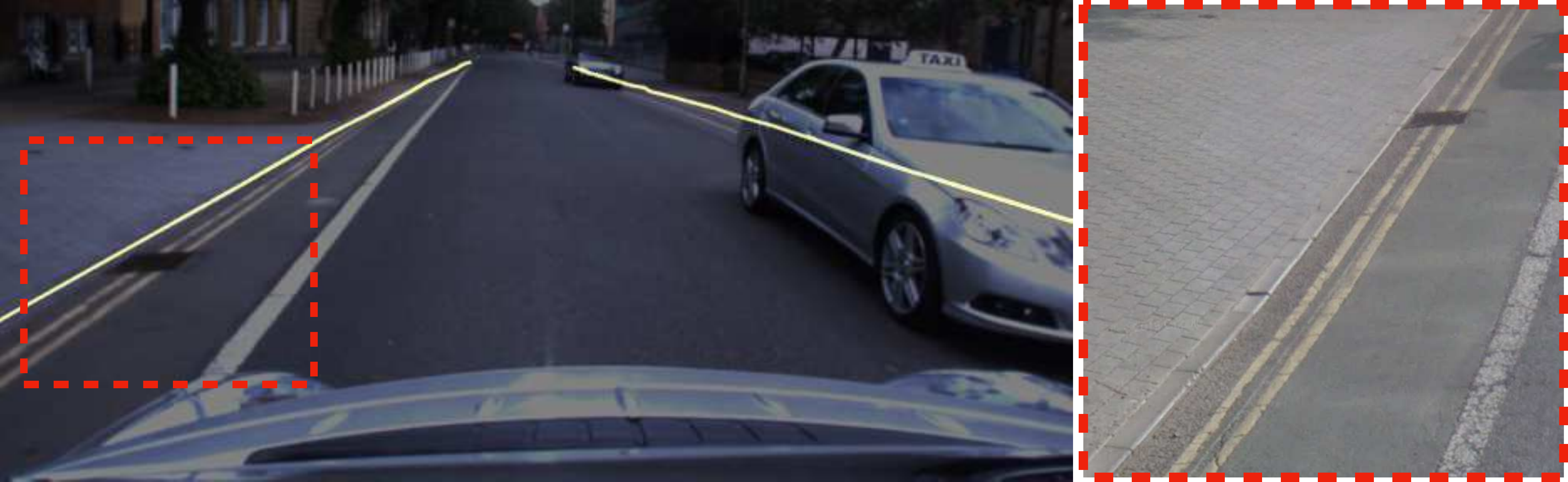}
\caption{The road boundary on the left hand side of the road with no height difference between road and pavement was precisely annotated in the laser pointcloud by simultaneously displaying its projection on the camera image and using it as a reference while annotating.}
\label{fig:non_curb_road_boundary_example}
\end{figure}

\review{
Annotation error in projecting from laser to camera due to offset between these two sensors is mitigated by hand-tuning the already fairly accurate extrinsic calibration provided by~\cite{RobotCarDatasetIJRR}.
Additionally, centimetre-level accuracy is available from our vision-based localiser.
Finally, drift in odometry is minimal over the limited distances over which estimates are accumulated.
}

\subsection{Extension to other traversals}
\label{sec:dataset:localiser}

\begin{figure}[!h]
    \centering
    \includegraphics[width=\columnwidth]{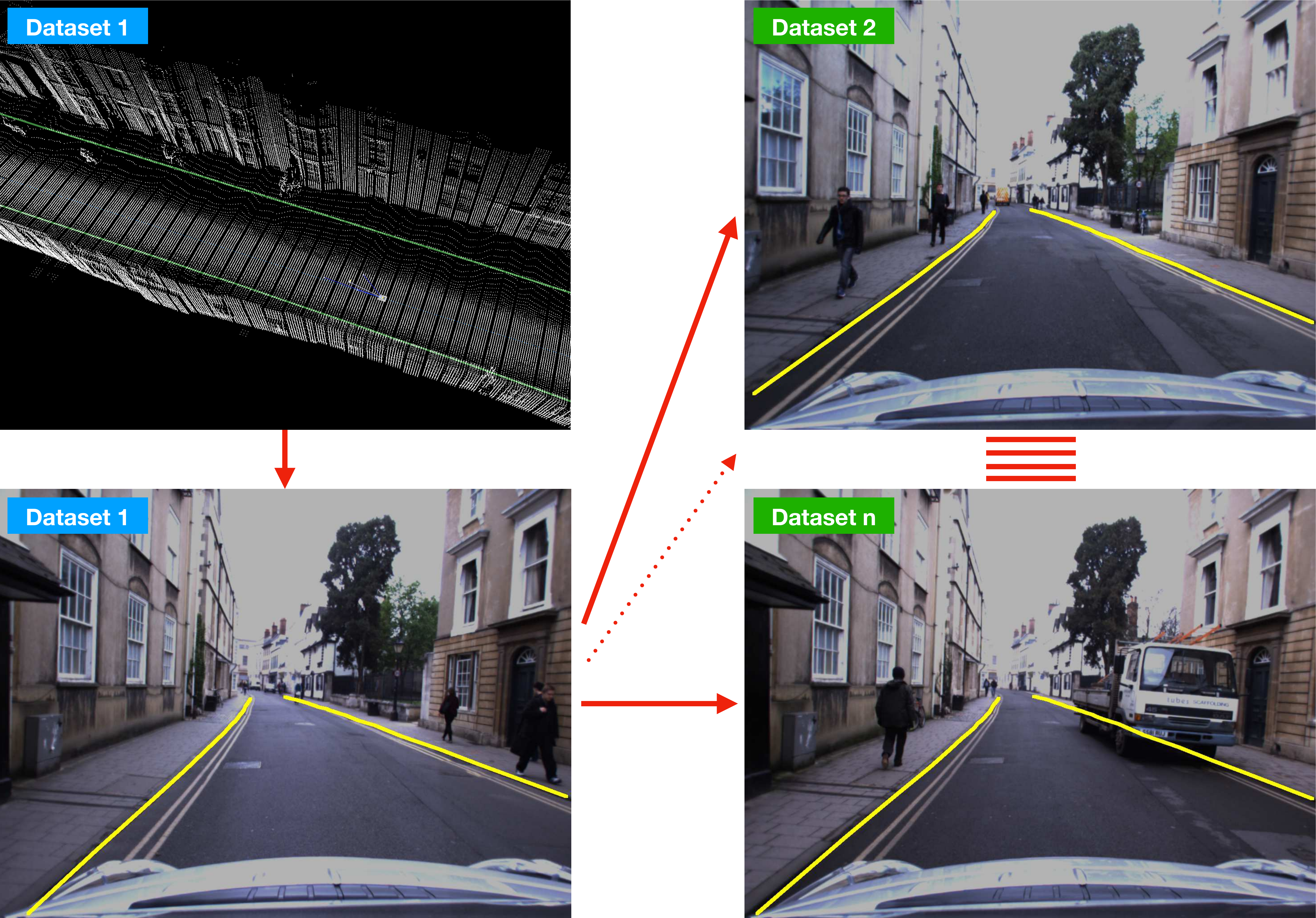}
    \caption{Boosting data samples: annotate a dataset once, project the annotations to the images of the same dataset, then run the training data generation tool using outputs from the vision-based classifier to automatically project the annotations to other traversals.}
    \label{fig:boosting_training_data}
\end{figure}

Annotating road boundaries in laser point clouds and projecting them into images enables us to easily generate hundreds of ground-truth masks within a short period of time, approximately \num{750} raw ground-truth masks per hour.
To boost the number of training samples, we used a vision-based localiser~\cite{churchill2013experience} to project labels from the annotated datasets to other traversals at different times and in different weather conditions, as shown in \cref{fig:boosting_training_data}.

When annotating road boundaries we did not assume that every road boundary was a curb and we annotated all road boundaries even if they were on a flat surface with no height differences between roads and pavements.
Although seeing flat road boundaries in the laser pointcloud was more difficult than curbs, we were simultaneously displaying projections of annotations of road boundaries on camera images in our annotation tool and using the projections as a reference to precisely position the annotations.
This makes the proposed road boundary annotation framework applicable to any type of road or road network (e.g., urban roads, dirt roads, highways and motorways).
An example of annotated road boundary with no height difference is shown in Figure \ref{fig:non_curb_road_boundary_example}.

\begin{figure}
\centering
\includegraphics[width=\columnwidth]{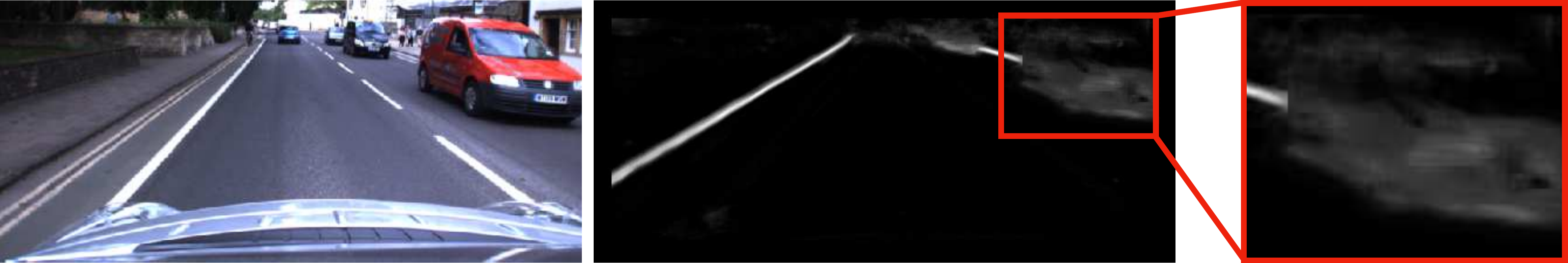}
\caption{
Blurry masks over occluding obstacles as understood by U-Net.
}
\label{fig:unet_partitioning}
\end{figure}

\subsection{Manual Curation}
\label{sec:dataset:curation}

We manually curated the raw masks of the dataset, removing sections where projected annotations were unreasonable, e.g. when localisation failed or where the location of road boundaries  changed over time.
As a result, we obtained \num{\totalUncurated} labelled samples, and from those, we edited \num{\totalCurated} samples while curating.
Note that every sample is a pair of images from the left and right lenses.

\subsection{Partitioning}
\label{sec:dataset:partitioning}

Having data samples separated in two classes, visible and occluded, can be beneficial for operational safety~\cite{Suleymanov2018Curb}.
To partition our raw road boundary masks into these two classes we trained the U-Net architecture \cite{RonnebergerFB15unet} with raw masks.
The resulting network can detect visible road boundaries, but fails to infer correct structure and position of occluded road boundaries for two reasons: (1) the network does not have a large enough receptive field to capture contextual information in the scene and to estimate position and structure of occluded road boundaries and (2) the network does not have any structure to force/bias outputs to be in a thin long shaped form for the occluded road boundaries.
As such, the network can detect visible road boundaries when trained with raw masks, but fails to infer occluded ones and outputs blurry masks over occluding obstacles as shown in \cref{fig:unet_partitioning}, enabling automatic partitioning of the raw training data into two classes. \review{We obtain masks for detected visible boundaries by applying threshold to the outputs. AND operation between the raw labels and thresholded outputs give us labels for visible road boundaries. Labels for occluded road boundaries are obtained by subtracting labels for visible from the raw labels} (see \cref{fig:main_figure} for examples of partitioned training data masks).
In total, 33\% of all labels and 32\% of curated labels are classified as occluded road boundaries.
\review{We refer the reader to~\cite{Suleymanov2018Curb} for more details on the performance of this network}.

\subsection{Recommended Usage}

Our suggested usage is that all samples can be used for pretraining and that fine-tuning with curated samples is sensible.
We used a subset of this dataset, approximately \num{23000} samples, for training our models in \cite{Suleymanov2018Curb} and achieved $F_1$ score of \num{0.9477} for visible road boundary detection and \num{0.8934} for occluded road boundary inference.
In this application, we dilated masks before training models.

The thickness of road boundary line annotations on the ground truth masks are the same regardless of the height of road boundaries.
This is because the annotated points do not contain any such information.
To compensate for this, we recommend to use tolerance-based evaluation metrics -- evaluation should be done by considering a sweep of thresholded pixel distances of predictions from true road boundaries.

\subsection{Release}

This section details the download format and tools related to this release.

The annotations are released alongside the associated RGB imagery as single \texttt{.tar} archives.
The following file structure details the data tree utilised for each traversal in the dataset.

\begin{lstlisting}
oxford-road-boundaries-dataset
|__ yyyy-mm-dd-HH-MM-SS
|   |__ uncurated
|   |   |__ raw.mp4
|   |   |__ classified.mp4
|   |   |__ left
|   |   |   |__ rgb
|   |   |   |__ raw mask
|   |   |   |__ classified mask
|   |   |__ right
|   |       |__ rgb
|   |       |__ raw mask
|   |       |__ classified mask
|   |
|   |__ curated
|       |__ raw.mp4
|       |__ classified.mp4
|       |__ left
|       |   |__ rgb
|       |   |__ raw mask
|       |   |__ classified mask
|       |__ right
|           |__ rgb
|           |__ raw mask
|           |__ classified mask
|__ ...
\end{lstlisting}

In total, we release 176 GB of data.
Moreover, the original \gls{sdk} of the \textit{Oxford RobotCar Dataset}\footnote{\urlstyle{tt}\rurl{github.com/ori-mrg/robotcar-dataset-sdk}} has been updated to include \texttt{Python} data loaders for the annotations for easy usage by the community.

\begin{figure*}
    \centering
    \includegraphics[width=0.245\textwidth]{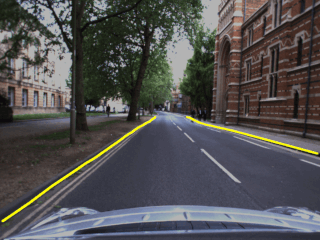} \hskip -0.5ex
    \includegraphics[width=0.245\textwidth]{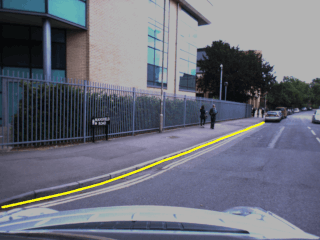} \hskip -0.5ex
    \includegraphics[width=0.245\textwidth]{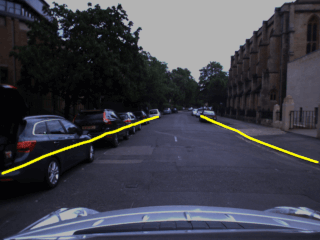} \hskip -0.5ex
    \includegraphics[width=0.245\textwidth]{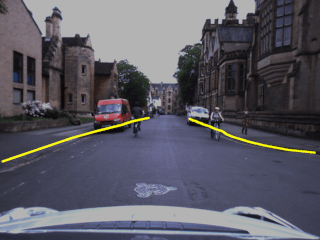}
    \vskip 0.3ex
    \includegraphics[width=0.245\textwidth]{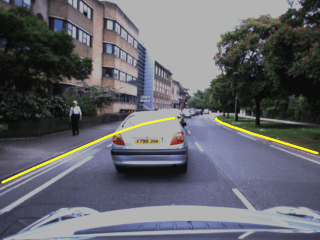} \hskip -0.5ex
    \includegraphics[width=0.245\textwidth]{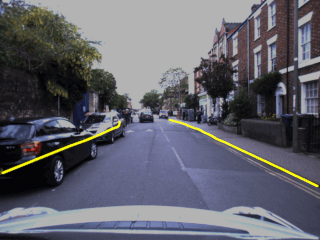} \hskip -0.5ex
    \includegraphics[width=0.245\textwidth]{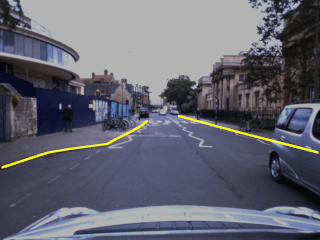} \hskip -0.5ex
    \includegraphics[width=0.245\textwidth]{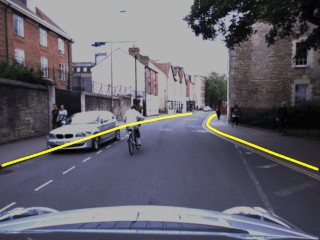}
    \vskip 0.3ex
    \includegraphics[width=0.245\textwidth]{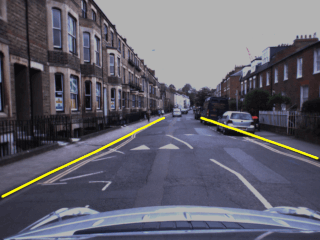} \hskip -0.5ex
    \includegraphics[width=0.245\textwidth]{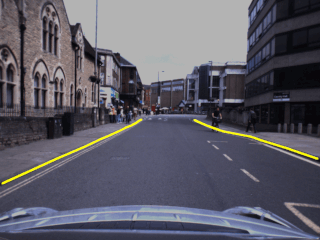} \hskip -0.5ex
    \includegraphics[width=0.245\textwidth]{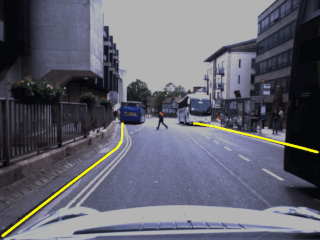} \hskip -0.5ex
    \includegraphics[width=0.245\textwidth]{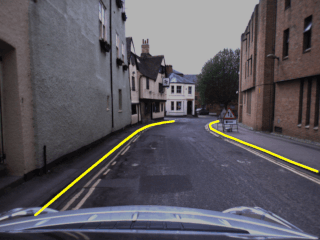}
    \caption{Generated ``raw'' road boundary examples: road boundary masks overlaid on top of RGB images. These masks are generated by projecting labels from annotated 3D point clouds into the corresponding images and they contain both visible and occluded road boundaries.}
    \label{fig:raw_road_boundaries}
\end{figure*}

\begin{figure*}
    \centering
    \includegraphics[width=0.245\textwidth]{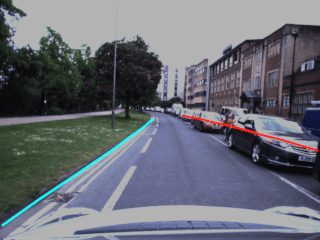} \hskip -0.5ex
    \includegraphics[width=0.245\textwidth]{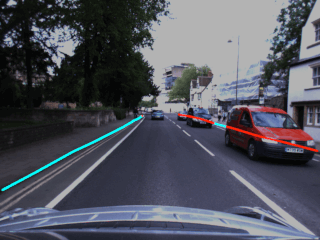} \hskip -0.5ex
    \includegraphics[width=0.245\textwidth]{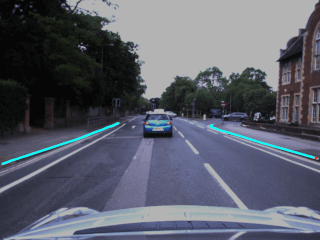} \hskip -0.5ex
    \includegraphics[width=0.245\textwidth]{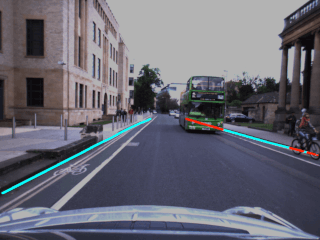} 
    \vskip 0.3ex
    \includegraphics[width=0.245\textwidth]{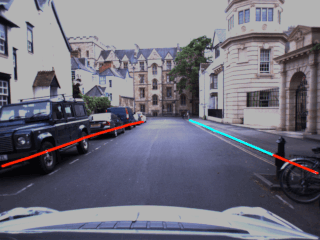} \hskip -0.5ex
    \includegraphics[width=0.245\textwidth]{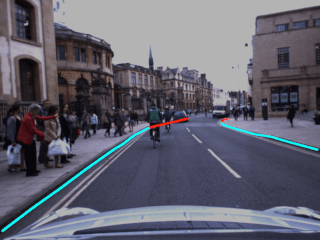} \hskip -0.5ex
    \includegraphics[width=0.245\textwidth]{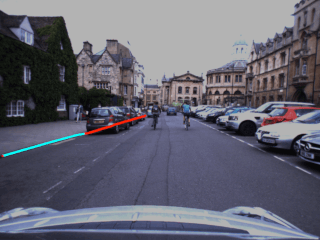} \hskip -0.5ex
    \includegraphics[width=0.245\textwidth]{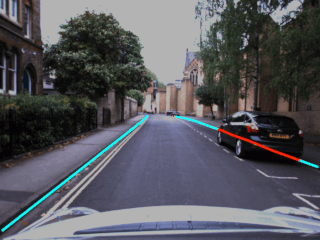}
    \vskip 0.3ex
    \includegraphics[width=0.245\textwidth]{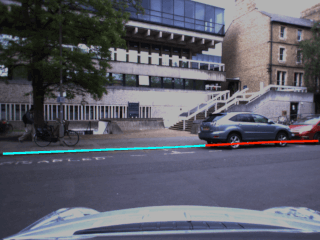} \hskip -0.5ex
    \includegraphics[width=0.245\textwidth]{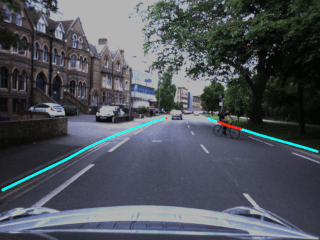} \hskip -0.5ex
    \includegraphics[width=0.245\textwidth]{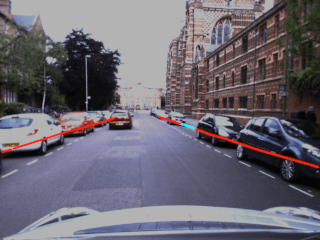} \hskip -0.5ex
    \includegraphics[width=0.245\textwidth]{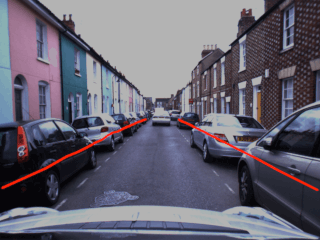}
    \caption{Partitioned road boundary data examples with two classes: visible (cyan) and occluded (red).}
    \label{fig:classified_road_boundaries}
\end{figure*}

\section{Conclusion and Future Work}
\label{sec:conclusion}

We presented a new autonomous driving dataset with a focus on road boundaries.
There is a scarcity of available datasets that include this annotation and resources which do exist do not provide pixel-wise locations for occluded road boundaries.
We have released \num{\totalUncurated} labelled samples, of which \num{\totalCurated} samples are curated.
This represents data from approximately \SI{70}{\kilo\metre} of driving.
We hope that this release proves useful to autonomous driving deep learning community.

In the future, we plan to complement this release with the laser-based lane boundary dataset presented in~\cite{Suleymanov2019Curb}. \review{Additionally, we plan to provide a benchmark for road boundary detection with a set of evaluation metrics to be used with the dataset.}

\section*{Acknowledgments}

\review{This work was presented at the workshop ``3D-DLAD: 3D-Deep Learning for Autonomous Driving'' (WS15), IV2021.}

The authors gratefully acknowledge support from the Assuring Autonomy International Programme, a partnership between Lloyd’s Register Foundation and the University of York, as well as EPSRC Programme Grant ``From Sensing to Collaboration'' (EP/V000748/1).

\bibliographystyle{IEEEbib.bst}
\bibliography{biblio}

\end{document}